\documentclass[letterpaper]{article} % DO NOT CHANGE THIS
\usepackage{aaai25}  % DO NOT CHANGE THIS
\usepackage{times}  % DO NOT CHANGE THIS
\usepackage{helvet}  % DO NOT CHANGE THIS
\usepackage{courier}  % DO NOT CHANGE THIS
\usepackage[hyphens]{url}  % DO NOT CHANGE THIS
\usepackage{graphicx} % DO NOT CHANGE THIS
\usepackage{natbib}  % DO NOT CHANGE THIS AND DO NOT ADD ANY OPTIONS TO IT
\usepackage{caption} % DO NOT CHANGE THIS AND DO NOT ADD ANY OPTIONS TO IT
\usepackage{algorithm}
\usepackage{algorithmic}
\usepackage{fancyvrb}

\usepackage{newfloat}
\usepackage{listings}
\DeclareCaptionStyle{ruled}{labelfont=normalfont,labelsep=colon,strut=off} % DO NOT CHANGE THIS
\floatstyle{ruled}
\newfloat{listing}{tb}{lst}{}
\floatname{listing}{Listing}
\nocopyright

\title{Which LLM is Best for Translating Natural Language Goals to PDDL}

\author{
Tom\'a\v{s} Balyo, Luk\'a\v{s} Chrpa, and G. Michael Youngblood
}
\affiliations{
Filuta AI, Inc., 1606 Headway Cir STE 9145, Austin, TX 78754, USA\\
\{tomas, lukas, michael\}@filuta.ai
}

\usepackage{bibentry}
\begin{document}

\maketitle

\begin{abstract}
Bridging the gap between human intent and machine execution remains a challenge in automated planning, where expressing goals in formal languages like PDDL restricts accessibility to non-experts. This paper empirically evaluates whether current Large Language Models (LLMs) can reliably translate natural language testing goals, written in informal language by video game testers, into well-formed PDDL targets suitable for classical planning. We present a carefully designed prompt template, integrating insights from iterative experimentation, aimed at maximizing both accuracy and response coherence from multiple state-of-the-art LLMs. Six contemporary models are systematically assessed on correctness, speed, and error tendencies using real-world, domain-specific benchmarks. All models demonstrate high correctness, exceeding 92\%, with Gemini 2.5 Flash achieving the highest accuracy at 96\% and the lowest incidence of false positives, while GPT-4.1 leads in response speed. Despite these advances, critical distinctions exist in model performance, and occasional failures arise from language ambiguity and limitations in domain representation. Our analysis underscores both the significant progress and ongoing gaps in enabling LLMs to act as robust bridges between natural language objectives and automated planning pipelines.
\end{abstract}


\section{Introduction}
Bridging the distance between human intention and machine execution stands as one of the challenges of Artificial Intelligence (AI). Nowhere is this divide more apparent than in the domain of automated planning, where expressing one’s objectives requires mastery of formal languages that are both powerful and unforgiving. For decades, systems like STRIPS and the Planning Domain Definition Language (PDDL) have enabled remarkable advances in planning and problem-solving, yet their benefits remain largely confined to those fluent in symbolic logic and domain engineering.

At the same time, we find ourselves in an era where Large Language Models (LLMs) are redefining the boundaries of machine understanding. These systems, trained on the full spectrum of human knowledge, adeptly translate between natural language and the structured requirements of a host of downstream applications. The goal of this paper is to examine the potential of letting humans express goals in plain language and then use LLMs to produce the well-formed inputs required by planning systems. Although the focus of this paper is only on the goal translation, it is an important element in making the planning technology accessible to a user who is not a planning expert. Arguably, LLMs are not yet capable of replacing planning experts in providing domain models~\cite{LLM-ICAPS2025}, on the other hand, a single domain model often can describe required game mechanics and thus be used for hundreds of tests. Initial states can be extracted directly from the game. Specifying each test goal by hand, however, might be a strenuous task for both the user and a planning expert, yet it might not be difficult for LLMs, especially if the domain model is provided within the prompt.

But potential is only as good as the performance in the wild. Translating goals for video game testing, where instructions are written informally and planning domains can be surprisingly nuanced, provides a fertile proving ground. This paper explores this practical challenge, evaluating the latest family of LLMs across multiple providers to determine their accuracy and speed. In doing so, we examine what it takes for LLMs to serve as reliable translators between human goals and their PDDL representations, and we shed light on which models are currently up to the task. This evaluation is grounded in the needs of real testers, real planners, and the persistent need to bring expressive AI systems to everyone, not just the experts.

\section{Related Work}\label{sec:related_work}

Exploiting LLMs in the field of Automated Planning has recently attracted a lot of attention~\cite{cao2025largelanguagemodelsplanning}. Plan generation via LLMs has been studied in the field of robotics, where LLMs are usually capable of generating valid plans, although often not very complex~\cite{zeng2023-llmrobotics, wang2024-llmrobotics}. In domain-independent settings, LLMs can be trained to solve problem instances sharing the same domain model, yet the scalability of the approach is limited (i.e., LLMs are not capable of generating valid plans for larger problem instances that are in the training set)~\cite{DBLP:conf/icaps/RossettiTGPSCO24}. In a traditional sense, LLMs are not capable of planning~\cite{kambhampati2024llms, goebel2025llmreasoningmodelsreplaceclassical} and, on top of that, it has been empirically shown that LLMs are not error-proof on computationally easy tasks associated with planning, such as identifying applicable actions or plan verification~\cite{DBLP:conf/aaai/Kokel0SS25}. 

Since LLMs struggle with plan generation, to leverage traditional symbolic planning, we have to provide domain and problem models. Arguably, LLMs can offer valuable assistance in the process of acquiring symbolic planning models from requirements provided in natural language~\cite{tantakoun2025llms, smirnov2024generatingconsistentpddldomains}. However, the use of LLMs in domain model acquisition (i.e., acquiring action schemes, predicates, or fluents representing the environment) still has some limitations that require the involvement of human experts in the process~\cite{LLM-ICAPS2025}. 

Although we still need an expert to obtain a (symbolic) domain model, the effort of the expert is usually needed only for developing the model and maintaining it (if a change in the requirements arrives). In the context of testing computer games, a domain model is developed once and modified only if there is a major change in the game~\cite{our-IJCAI}. However, a planning task description has to be generated for every test. LLMs can be used to extract a full specification of a planning task~\cite{liu2023llmpempoweringlargelanguage,agarwal2024tictranslateinfercompileaccuratetext, zuo2025planetariumrigorousbenchmarktranslating}. However, this is not necessary for game testing as the initial state can be extracted from the game. Specifying goals by users, on the other hand, remains the bottleneck. In literature, there also exist specialized approaches that only focus on acquiring symbolic goal specification from a user's text in natural language~\cite{xie2023translatingnaturallanguageplanning,lyu2023faithfulchainofthoughtreasoning}. For more details about the existing techniques, the interested reader is referred to~\cite{tantakoun2025llms}. Our work, albeit sharing similar characteristics (e.g., including PDDL domain model into the prompt), also focuses on providing feedback on \emph{negative cases} in which user-specified goals cannot be expressed as a PDDL goal for the given domain model. 

\section{Preliminaries}
\label{sec:preliminaries}

This section introduces fundamental concepts in automated planning and Large Language Models (LLMs) necessary for understanding this paper.

\subsection{Automated Planning}
\textit{Automated Planning} is an area of Artificial Intelligence concerned with the development of algorithms that generate sequences of actions to achieve a goal from an initial state. We primarily focus on \textbf{Classical Automated Planning}, which operates under several key assumptions: the world is static (no exogenous events), deterministic (actions have predictable outcomes), fully observable, and discrete.

A widely adopted framework for classical planning is \textbf{STRIPS} (STanford Research Institute Problem Solver) \cite{fikes1971strips}. In STRIPS, the world state is defined by a set of logical propositions. Actions have preconditions (propositions that must be true for the action to be executable) and effects (propositions that become true or false after the action).

\textbf{Fluents} are propositions or predicates that can change their truth value over time as actions are executed. They represent the dynamic aspects of the world state (e.g., \texttt{(robot-at roomA)}).

The \textbf{Planning Domain Definition Language (PDDL)} \cite{mcdermott1998pddl} is the de facto standard language for representing planning problems. A PDDL planning problem is split into two files:
\begin{itemize}
    \item A \textbf{domain file}: This defines the types of objects, predicates (fluents), and actions that can be performed within a specific planning environment. It describes the general rules and capabilities.
    \item A \textbf{problem file}: This specifies a particular instance of the planning problem for a given domain, including the specific objects, their initial state (initial assignments of fluents), and the desired goal state.
\end{itemize}

\subsection{Large Language Models (LLMs)}
\textbf{Large Language Models (LLMs)} are a class of artificial intelligence models, typically based on the transformer architecture \cite{vaswani2017attention}, trained on vast amounts of text data to understand, generate, and process human language. Their strength lies in their ability to perform a wide range of natural language understanding and generation tasks, often through few-shot or zero-shot learning, as exemplified by models like GPT-3 \cite{brown2020language}.

A \textbf{prompt template} is a structured text input given to an LLM, designed to guide its response towards a specific task or format. It typically contains placeholders that are filled with task-specific information (e.g., instructions, context, examples, or, in this paper, planning problem components).

\begin{figure*}
\begin{Verbatim}[frame=single]
   
    The task is to translate test goals expressed in a natural language 
    text (written by someone who writes video game testing instructions)
    to automated planning goals in the PDDL language. 
    We are dealing with classical STRIPS planning with numeric fluents.
    
    As input I will provide you with the domain in PDDL format, the 
    list of all available objects along with their types 
    and finally the natural language text representing the test goals.

    It may be the case that the provided natural language goals cannot 
    be fully expressed as PDDL goals. Possible reasons for this could be:
    - the goal cannot be expressed in the formalism of classical planning
      with numeric fluents in general
    - the provided domain does not contain predicates or function 
      definitions that match the desired goals
    - the goal cannot be expressed with the available objects provided
    - the provided goals are too ambiguous
    
    If the natural language goal is ambiguous or cannot be fully represented
    by the provided domain and objects, provide a short (2 or 3 sentences) 
    explanation. In this this case start the message with 
    "Goal cannot be expressed".
    
    If the goal can be expressed, then answer with the goal specification
    in PDDL format (the "(:goal)" clause) and  nothing else (no explanation, 
    no comments, no remarks, no header or title). Try to express the goals 
    simply and succinctly, i.e., avoid quantifiers and use as few expressions
    and predicates as necessary. Here is an example response:
    
    (:goal
        (and
            (made p1 pulse_tank_type)
            (<= 1000 (resources p1))
        )
    )

    Now I provide the actual input:

    <domain>
    {domain}
    </domain>
    <objects>
    {objects}
    </objects>
    <goal>
    {human_goal}
    </goal>
\end{Verbatim}
\caption{The complete prompt template used for evaluating all six Large Language Models (LLMs). The placeholder \texttt{\{domain\}} is replaced by the contents of domain.pddl, \texttt{\{objects\}} by a list of all objects and their types, and \texttt{\{human\_goal\}} by the desired goal provided in natural language.}
\label{fig_prompt}
\end{figure*}

\section{Prompt Template Design}
\label{sec:prompt_design}

The complete prompt template utilized in our evaluation is presented in Figure~\ref{fig_prompt}. This specific formulation was developed through an iterative refinement process, incorporating empirical insights gained from initial experimental trials and addressing observed limitations.

Initially, our prompt included the full contents of the \texttt{problem.pddl} file, rather than a concise list of objects and their types. However, we observed a significant decrease in LLM response accuracy for benchmarks characterized by large problem files, specifically those with a high number of initial state predicates. We hypothesize that this degradation in performance is attributable to the total prompt length exceeding a reasonable context window. This phenomenon, where models struggle to effectively utilize or prioritize information in very lengthy inputs, is known to cause performance degradation with long contexts, sometimes referred to as ``lost in the middle'' \cite{liu2023lost, hong2024context}, and relates to the inherent complexities of the Transformer architecture \cite{vaswani2017attention}. Consequently, the revised template incorporates only the essential object definitions to minimize overall prompt length.

Furthermore, an earlier iteration of the prompt frequently led to a high incidence of false positive answers. In these cases, the LLMs generated PDDL goals even when the natural language objective could not be properly expressed or was unachievable within the domain's constraints. To mitigate this over-generation of invalid goals, we integrated specific examples within the prompt. These examples demonstrate various reasons why certain types of natural language goals might be inexpressible in PDDL (e.g., requiring new predicates not defined in the domain, or necessitating actions not available). This guided the LLMs towards more constrained and accurate responses, reducing the propensity for false positives.

Finally, we observed that some LLM-generated PDDL goals were not only overly complex and convoluted but also frequently incorrect. Such intricate and erroneous formulations posed significant challenges for subsequent automated planning stages. To address this issue, a clear directive requesting simple and concise answers was integrated into the prompt. This modification proved highly effective, as it led the LLMs to more consistently produce correct PDDL goals, while simultaneously ensuring their optimal formulation and interpretability for practical use within the planning pipeline.

\section{Experimental Evaluation}
We are addressing a task that uses LLMs to translate natural language goals into a formal planning language. The inputs for this task are:
\begin{itemize}
    \item A planning domain model in PDDL, with features typical of the numeric tracks in international planning competitions.
    \item A planning problem instance, which includes objects and an initial state but has an empty goal.
    \item A planning goal expressed as unstructured natural language text.
\end{itemize}
The expected output depends on whether the given natural language goal can be represented within the provided domain and problem instance:
\begin{itemize}
    \item \textbf{If the goal is expressible:} The output should be a valid PDDL expression representing the goal.
    \item \textbf{If the goal is not expressible:} The output should be a brief, natural language explanation (a few sentences) of why the goal cannot be modeled with the given domain and problem.
\end{itemize}

\begin{figure*}
\centering
\includegraphics[width=1.9\columnwidth]{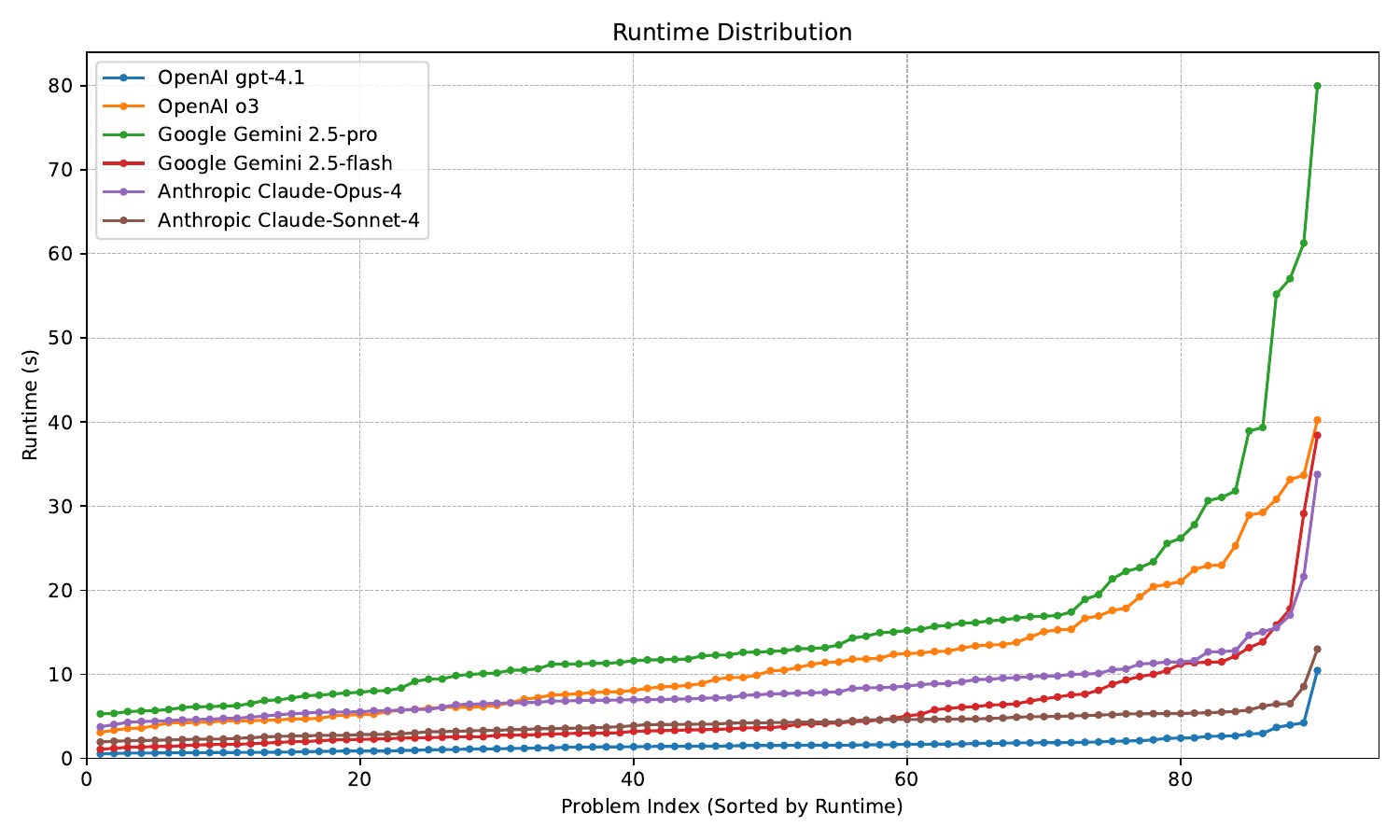} 
\caption{Comparative Prompt Response Time Analysis of Large Language Models (LLMs). Each line represents a distinct LLM, plotting its prompt response time (y-axis) against the problem index (x-axis) after sorting all prompts by increasing runtime for that model.}
\label{fig1}
\end{figure*}

\subsection{LLM Usage}
For this evaluation, we selected two models from each of three leading LLM providers: OpenAI, Google, and Anthropic. From each provider, we chose their flagship model alongside their performance-optimized ``lightweight'' or ``fast'' model. All models were accessed via their standard APIs, and each was evaluated using a single prompt for every benchmark that is based on the identical template described above. The used LLMs are the following:
\begin{itemize}
\item OpenAI O3
\item OpenAI GPT 4.1
\item Google Gemini 2.5 Pro
\item Google Gemini 2.5 Flash
\item Anthropic Claude Opus 4
\item Anthropic Claude Sonnet 4
\end{itemize}

\subsection{Benchmarks}
Our benchmark suite comprises a total of 90 natural language goals. This set is balanced, with 45 goals that can be formally expressed in PDDL and 45 that cannot.

To ensure a diverse evaluation, the goals were created across eight distinct domains. Four of these are standard domains from the International Planning Competition (IPC): Rovers, Barman, Woodworking, and Parking. The remaining four domains were specifically designed to simulate real-world problem-solving scenarios, drawing inspiration from video games, including two real-time strategy games, one first-person shooter, and a city-building adventure. 
The natural language goals are displayed in the appendix of this paper in Tables \ref{table_lyra}, \ref{table_ipc} and \ref{table_silica}.

The natural language goal formulations were crowd-sourced from a diverse group of employees within our company. A significant majority of these goals (more than half) were written by individuals with no prior experience in PDDL modeling, and crucially, without access to the PDDL domain models. This approach was taken to ensure that the goals represent genuine human problem descriptions, free from the biases of a formal planning language.

\subsection{Results}
Our evaluation methodology utilized a semi-automated process. For the expressible goals, we established a set of reference PDDL goals. If an LLM's output matched one of these references exactly, it was automatically marked as correct. In all other cases, a PDDL expert conducted a manual review to determine if the generated goal was valid and semantically correct.

Conversely, for the goals that were inexpressible, the evaluation was automated. An LLM was considered to have answered correctly if its response indicated that the goal could not be modeled. It is important to note that we did not perform a manual check on the quality or detail of the natural language explanations provided for these cases.

To quantify the errors, we defined two metrics: A False Negative was recorded when an LLM failed to provide a PDDL expression for a goal that was actually expressible. Conversely, a False Positive was recorded when an LLM provided a PDDL goal for a problem that was fundamentally inexpressible.

\begin{table}
\centering
\begin{tabular}{l||r|r|l|l|l}
Model Name & correct & time & FP & FN & IS\\
\hline
OpenAI GPT-4.1 & 0.92 & \textbf{1.61} & 4 & 2 & \textbf{0} \\
OpenAI o3 & 0.93 & 11.49 & 3 & 2 & \textbf{0} \\
Gemini 2.5-pro & 0.94 & 15.83 & 3 & \textbf{0} & \textbf{0} \\
Gemini 2.5-flash & \textbf{0.96} & 5.49 & \textbf{2} & 1 & \textbf{0} \\
Claude-Opus-4 & 0.94 & 8.29 & 3 & 1 & 1 \\
Claude-Sonnet-4 & 0.93 & 4.10 & 3 & 3 & \textbf{0} \\
\end{tabular}
\caption{Summary of experimental results per LLM showing correctness rate, average prompt response time, number of false positives (FP), number of false negatives (FN), and responses containing invalid syntax (IS).}
\label{table1}
\end{table}

Table~\ref{table1} summarizes the experimental results, demonstrating that all six models achieved high correctness, exceeding 92\%. Gemini 2.5 Flash exhibited the highest correctness rate at 96\% and yielded the fewest false positives (2). Notably, Gemini 2.5 Flash surpassed its Pro counterpart (Gemini 2.5 Pro) in accuracy while maintaining significantly faster response times. Conversely, GPT-4.1 emerged as the fastest model, albeit with the lowest accuracy among those tested. A more granular comparison of model runtimes is presented in Figure~\ref{fig1}. Furthermore, only a single response with invalid syntax was generated across all evaluations.

Variation in LLM output does exist within (intra) and between (inter) models. Intra-model differences from repeated runs on the same LLM—arise largely from batch-size–dependent numerical effects and non–batch-invariant kernel operations, so even temperature-zero inference can yield diverging results~\cite{he2025nondeterminism}. Inter-model differences, by contrast, stem from deeper divergences in architecture, training data, and engineering choices, which shape how each system learns language, encodes context, and responds to ambiguity~\cite{liu2025understanding}. Together, these factors mean that LLM output variation is influenced both by subtle computational implementation and by broad design philosophy. 

\section{Conclusion}
This paper asks the question: can LLMs serve as bridges between natural human intent and the rigor of automated planning languages like PDDL? Our examination of LLM performance reveals both the significant improvements made and the limitations that remain. By systematically evaluating leading LLMs on the pragmatic challenge of translating informal video game test goals into executable PDDL targets, we have gained a realistic measure of where these technologies succeed and remain challenged.

The results are both encouraging and insightful. Current LLMs demonstrate a capability for this language translation by routinely delivering high correctness across complex, domain-specific scenarios. Not all models are created equal though. Our benchmarks highlight critical differences in accuracy, response times, and the frequency of subtle but important errors such as false positives or syntactic errors. These distinctions matter for practitioners who must choose the right tools for their pipelines. Our evaluation also surfaces that the distance between human expression and machine understanding, though narrowing, has not yet been closed. Some failures stem from the inherent ambiguity of language; others from the current limitations of LLMs in parsing nuanced domain logic, or managing edge cases where the formalism cannot keep pace with human creativity. Future work must focus on enhancing the interactive capabilities of these systems, enabling richer dialogue, clarification, and error recovery. The next steps are 1) expanding the breadth and depth of evaluation datasets, 2) refining prompt engineering practices, and 3) integrating feedback loops that better allow LLMs to handle real-world imperfections. 

Our findings provide a snapshot of the current state of the art and a practical pathway for harnessing the power of LLMs to unlock the value of automated planning for a broader audience. As this technology continues to advance, narrowing the distance between what we want to say and what machines can truly understand remains not just a goal, but a promise within reach.

\bibliography{aaai25}
%\newpage
\appendix
\section{Appendix: Natural Language Goals}

The specifications of the natural language goals we used in our experiments are provided in Tables~2--4.

\begin{table*}
\centering
\begin{tabular}{p{\linewidth}}
\textbf{Natural Language Goals (Translatable)}\\
\hline
collect each type of weapon available on the map\\
Test that healing works by finding and using a healthkit\\
Test that you can shoot from multiple weapons in 1 game instance\\
test that I can shoot directly with a pistol\\
Make sure the player has the gun, rifle, and gatling in their inventory. Also, the ammo count inside the containers (magazines) of the gun, rifle, and gatling must be empty (set to 0).\\
Restore the player's health to maximum.\\
Fire the gun at least once, then reload it.\\
Eliminate bot b1.\\
Reduce the gun’s total ammo to zero.\\
\\
\textbf{Natural Language Goals (Not translatable)}\\
\hline
Eliminate a bot with a shotgun\\
Eliminate two bots within a 10 seconds time interval\\
perform 3 headshots in a row\\
hide under a staircase\\
jump ten times\\
Verify that shooting a bot decreases his health points\\
Go through a portal\\
Validate that going through the portal won't reduce health\\
Validate you have all the weapons when going through the portal\\
verify shooting reduces ammo by 1\\
Verify you can't shoot when there is no ammo\\
test that crouching and leaning while shooting work\\

\end{tabular}
\caption{Natural language goals in the ``first person shooter'' domain. In the domain we model the weapons, the amount of ammunition and health of the player and bots.}
\label{table_lyra}
\end{table*}

\begin{table*}
\centering
\begin{tabular}{p{\linewidth}}
\textbf{barman} \\
\hline
Shot 1 must contain cocktail 4 shot 2 must contain cocktail 2 shot 3 must contain cocktail 3 and shot 4 must contain cocktail 1.\\
Shot 1 contains cocktail 2, shot 2 contains cocktail 3, shot 3 contains cocktail 4, and shot 4 contains cocktail 1.\\
Ensure that shot1 contains cocktail2, shot2 contains cocktail4, shot3 contains cocktail1, and shot4 contains cocktail3.\\
Shot 1 must contain cocktail 4.  
Shot 2 must contain cocktail 2.  
Shot 3 must contain cocktail 1.  
Shot 4 must contain cocktail 5.  
Shot 5 must contain cocktail 3.\\
\\
\textbf{rovers} \\
\hline
Successfully transmit the soil data collected at waypoint2, the rock data collected at waypoint3, and a high resolution image of objective1.\\
Transmit the soil data collected at waypoint0, the rock data collected at waypoint0, and the low-resolution image data of objective1.\\
Transmit soil data from waypoint2, transmit rock data from waypoint0, and transmit a color image of objective0.\\
\\
\textbf{woodworking} \\
\hline
Make sure that part p0 is available, has a smooth surface, and is varnished.  
Make sure that part p1 is available, made of teak wood, has a smooth surface, and is varnished.  
Make sure that part p2 is available, is coloured green, and has a smooth surface.  
Make sure that part p3 is available, made of mahogany wood, and has a smooth surface.  
Make sure that part p4 is available, made of teak wood, and is glazed.\\
Make sure you have all four parts available.  
- p0 must be mauve in color and have a smooth surface.  
- p1 must be blue in color and have a smooth surface.  
- p2 must be white in color, made of oak, have a smooth surface, and be glazed.  
- p3 must be mauve in color, made of pine, and be glazed.\\
Make sure at the end of your test that the following conditions are true:

- The part p0 is available, has the mauve colour, and its surface is smooth.
- The part p1 is available, has the green colour, is made of mahogany, its surface is very smooth, and it is varnished.
- The part p2 is available, has the mauve colour, and is glazed.
- The part p3 is available, has the white colour, is made of mahogany, its surface is smooth, and it is glazed.
- The part p4 is available, is made of teak, and is varnished.\\
Make sure the following objectives are completed:
- Part p0 is available, colored white, and has the glazed treatment.
- Part p1 is available, made of pine wood, and has the varnished treatment.
- Part p2 is available, colored blue, has a very smooth surface, and has the glazed treatment.
- Part p3 is available, colored black, and has a very smooth surface.
- Part p4 is available, has a very smooth surface, and is varnished.\\
\\
\textbf{parking} \\
\hline
Park car\_00 at curb\_0 with car\_07 directly behind it. Park car\_01 at curb\_1 with car\_08 directly behind it. Park car\_02 at curb\_2 with car\_09 directly behind it. Park car\_03 at curb\_3 with car\_10 directly behind it. Park car\_04 at curb\_4 with car\_11 directly behind it. Park car\_05 at curb\_5. Park car\_06 at curb\_6.\\
Park the cars so that curb\_0 has car\_00 at the curb with car\_07 directly behind it, curb\_1 has car\_01 at the curb with car\_08 directly behind it, curb\_2 has car\_02 at the curb with car\_09 directly behind it, curb\_3 has car\_03 at the curb with car\_10 directly behind it, curb\_4 has car\_04 at the curb with car\_11 directly behind it, curb\_5 has car\_05 at the curb, and curb\_6 has car\_06 at the curb.\\
Park car\_00 at curb\_0, with car\_07 behind car\_00. Park car\_01 at curb\_1, with car\_08 behind car\_01. Park car\_02 at curb\_2, with car\_09 behind car\_02. Park car\_03 at curb\_3, with car\_10 behind car\_03. Park car\_04 at curb\_4, with car\_11 behind car\_04. Park car\_05 at curb\_5. Park car\_06 at curb\_6.\\
Park car\_00 at curb\_0 with car\_07 directly behind it. Park car\_01 at curb\_1 with car\_08 directly behind it. Park car\_02 at curb\_2 with car\_09 directly behind it. Park car\_03 at curb\_3 with car\_10 directly behind it. Park car\_04 at curb\_4 with car\_11 directly behind it. Park car\_05 at curb\_5. Park car\_06 at curb\_6.\\
\end{tabular}
\caption{Natural language goals in the IPC domains. All translatable.}
\label{table_ipc}
\end{table*}

\begin{table*}
\centering
\begin{tabular}{p{\linewidth}}
\textbf{Natural Language Goals (Translatable)}\\
\hline
build a pulse tank and use it to destroy an enemy light quad\\
as player 1 reach research level 4\\
Test that it's possible to build air fighter unit\\
Test that it's possible to build all buildings\\
Reach highest possible research level\\
Collect 40000 resources as fast as possible\\
Check that enemy Silo can be destroyed\\
Player 1 must build at least one Pulse Tank. Player 2 must build at least one Rifleman. Then, Player 1 must destroy at least one of Player 2’s Riflemen using a Pulse Tank.\\
Make sure that by the end of the test, Player 1 has successfully produced at least one Hover Tank and at least one Rifleman unit.\\
Have player 1 build a Rocket Launcher and a Radar, and have player 2 train a Rifleman and construct a Light Vehicle Factory.\\
The goal is to ensure that player 1 has successfully produced at least one Pulse Tank unit.\\
Build a Radar and a Silo as player 1.\\\\
\textbf{Natural Language Goals (Not translatable)}\\
\hline
harvest 10000 resources within 20 minutes of gameplay\\
build a rifleman and have it destroy your own harvester\\
destroy enemy's base within 30 minutes\\
check that eliminating a tank with rifleman takes longer than with a rocket launcher\\
check that you can drive over infantry units with vehicle, effect: infantry driven over is eliminated\\
destroy your own infantry production building\\
test that harvester returns automatically if fired upon\\
test that the more units the more space they have to occupy (not 100 soldiers at 1 square meter)\\
test that building a soldier on higher tech level is faster\\
test that when approaching enemy base, you are spotted earlier when approaching over desert than over mountains\\
test that you can start building during harvester unloading, i.e., don't have to wait till harvester is empty, but when I have enough resources, I can build\\
test parallel attack of two friendly armies against one CPU enemy (aka multiplayer game)\\
test that keyboard shortcuts work as set in options, i.e. don't have to click on units, or "b" moves my focus on base, "h" moves focus on harvester, etc\\
test the shortest strategy to defeat a CPU opponent on map 1\\
keep the game playing for 60 minutes in a row\\
check that turrets around your base can stop the CPU opponent better than infantry (1:1 counts of turrets:infantry)\\
scout through map and find unreachable places, plot map of places where a unit can not go\\
check that speed of a light armored vehicle is greater than a heavy armored tank\\
test that a light vehicle can reach places which other heavy units (tanks) can't; or similar scenario, comparing pairs of units against each other in terms of speed, way to cross terrains, fire cadence)\\
test that building two units in two separate factories takes approx. the same time\\
does the game crash or misbehave with 1000 units on map?\\
shooting, aiming, crouching, sprinting works across all weapon types - each unit can move, can fire, can return to base\\
test coordinated attack of multiple units against an enemy base\\
unit pathfinding follows terrain and obstacles as expected, test that units do not get stuck during long gameplay\\
building construction works across all valid terrain types, can't build on rock, on sand, on water\\
saved games are loaded correctly, units, resources, and game state are restored as expected.\\
test that automated base defenses react properly to enemies\\
build an army of 5 tanks (deliberately ambiguous term 'tank')\\
A rifleman shooting at other rifleman should win the shootout in 50\% times\\
Verify you can't build two buildings at the same place\\
Verify you can't build research level 8 before building research level 6\\
Unit is destroyed when it reaches 0 hitpoints\\
Building is destroyed when it reaches 0 hitpoints\\
\end{tabular}
\caption{Natural language goals in the ``real time strategy'' domain. In the domain we model gathering resources, doing research and constructing buildings and units.}
\label{table_silica}
\end{table*}

\end{document}